%% file: main.tex
\documentclass[11pt,a4paper]{rsi_report}

\usepackage{booktabs}
\usepackage{array}
\usepackage{longtable}
\usepackage{graphicx}
\usepackage{fontawesome5}
\usepackage{float}
\usepackage{afterpage}
\usepackage{algorithm}
\usepackage{algpseudocode}
\usepackage{xcolor}
\usepackage{colortbl}
\usepackage{xfp}
\usepackage{hhline}
\usepackage{hyperref}
\usepackage{url}
\usepackage[most]{tcolorbox}
\usepackage[normalem]{ulem}
\let\savedttdefault\ttdefault
\usepackage{inconsolata}
\let\ttdefault\savedttdefault
\usepackage{needspace}
\usepackage{enumitem}

\definecolor{reportLink}{HTML}{1A73E8}
\definecolor{reportCite}{HTML}{174EA6}
\hypersetup{colorlinks=true,linkcolor=reportLink,citecolor=reportCite,urlcolor=reportLink}

\newcommand{\rsifigref}[1]{\hyperref[#1]{\textcolor{blue}{Figure~\ref*{#1}}}}
\newcommand{\rsienv}{\textsc{RSI-Env}}

\title{\textsc{RSI-Env}: An Open-Ended Environment for Recursive Self-Improvement of Agents}

\author[1,*]{Wanying Qu}
\author[2,*]{Zhiyuan Fan}
\author[1]{Jiyao Liu}
\author[3]{Jiang Zhou}
\author[4,5]{Jianghan Shen}
\author[1,6,\textdagger]{Yanwei Fu}
\author[5,\textdagger]{Dongrui Liu}
\affil[1]{Fudan University}
\affil[2]{The Hong Kong University of Science and Technology}
\affil[3]{Georgia Institute of Technology}
\affil[4]{Nanjing University}
\affil[5]{Shanghai Artificial Intelligence Laboratory}
\affil[6]{Shanghai Innovation Institute}
\renewcommand{\reportinstitutionlogos}{%
  \makebox[\textwidth][s]{%
    \raisebox{-0.5\height}{\includegraphics[height=32pt,
      trim=0bp 221.52bp 366.96bp 0bp,clip]{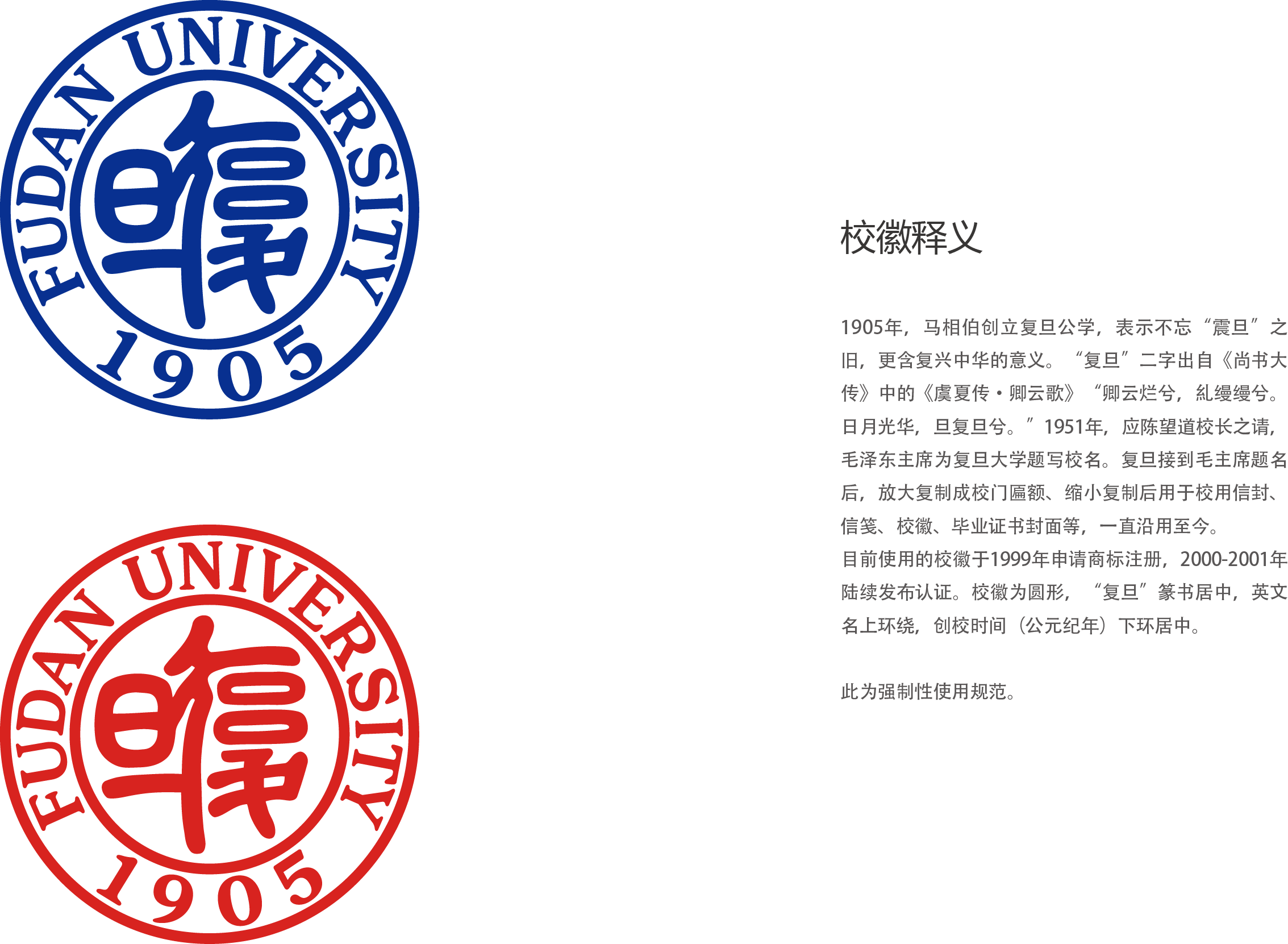}}\hfill
    \raisebox{-0.5\height}{\includegraphics[width=90pt,height=32pt,
      keepaspectratio]{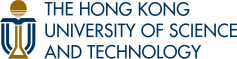}}\hfill
    \raisebox{-0.5\height}{\includegraphics[width=80pt,height=32pt,
      keepaspectratio]{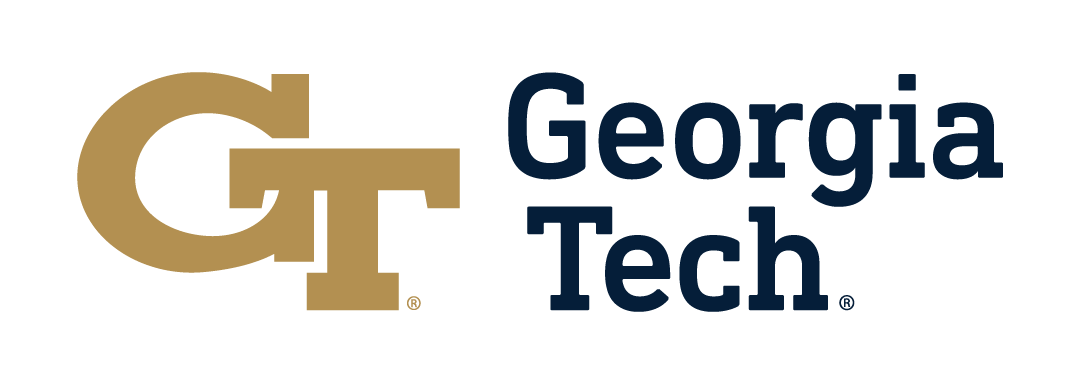}}\hfill
    \raisebox{-0.5\height}{\includegraphics[width=84pt,height=32pt,
      keepaspectratio]{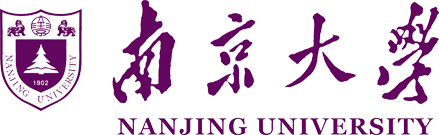}}\hfill
    \raisebox{-0.5\height}{\includegraphics[width=62pt,height=32pt,
      keepaspectratio]{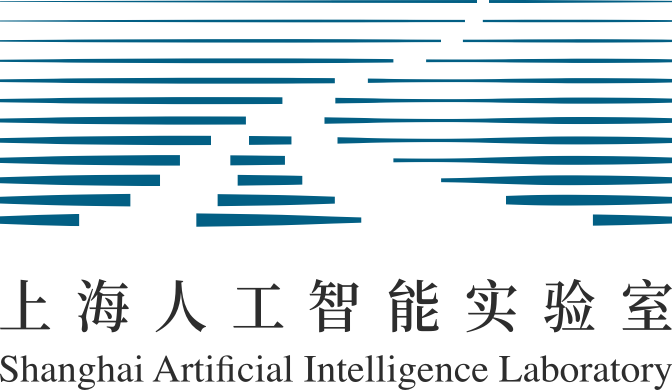}}\hfill
    \raisebox{-0.5\height}{\includegraphics[width=92pt,height=32pt,
      keepaspectratio]{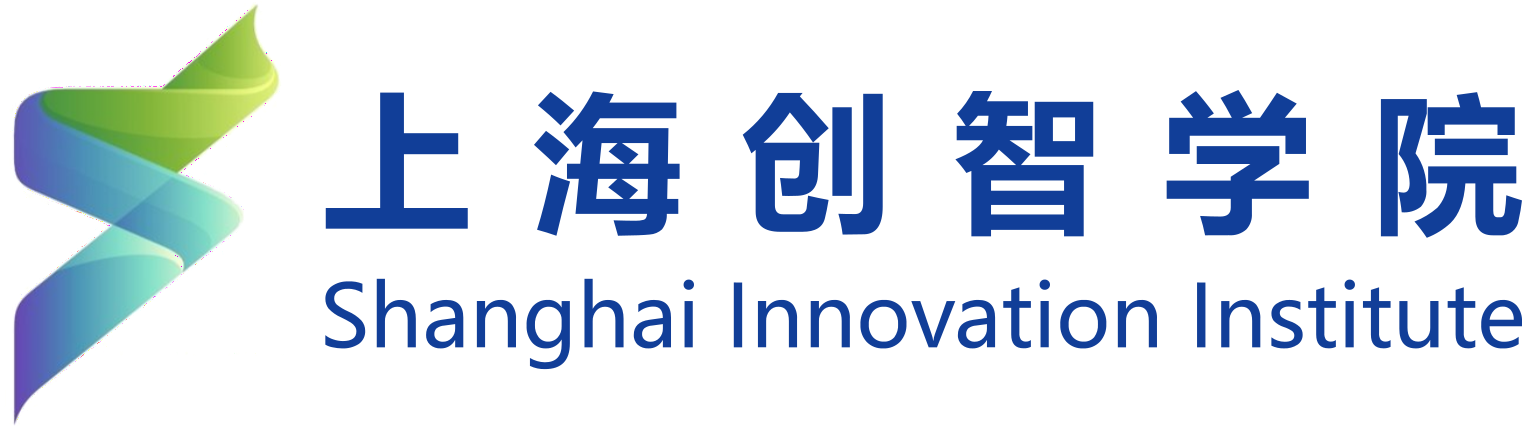}}%
  }%
}
\renewcommand{\reportauthornote}{%
  \parbox[t]{0.94\textwidth}{\reportfooterfont\raggedright
    \textsuperscript{*}Equal contribution\quad
    \texttt{wyqu24@m.fudan.edu.cn}, \texttt{zhiyuan.fan@connect.ust.hk}\\
    \textsuperscript{\textdagger}Corresponding authors%
  }%
}
\renewcommand{\reportprojectlinks}{}
\hypersetup{pdfauthor={Wanying Qu, Zhiyuan Fan, Jiyao Liu, Jiang Zhou, Jianghan Shen, Yanwei Fu, Dongrui Liu}}

\begin{document}

\maketitle

\afterpage{\afterpage{\clearpage\input{section/related_comparison}}}

\input{section/abstract}

\input{section/introduction}

\clearpage
\input{section/related_work}

\newpage
\input{section/method}

\input{section/experiments}

\input{section/conclusion}

\input{main.bbl}
\clearpage
\appendix
\input{section/appendix_contents}

\input{section/appendix_settings}

\input{section/appendix_prompts}

\end{document}

%% file: section/related_comparison.tex
\begin{table}[H]
\centering
\caption{Comparison of reported agent self-improvement settings. Self as Improver denotes updated systems conducting subsequent research.}
\label{tab:related-comparison}
\fontsize{8}{10}\selectfont
\setlength{\tabcolsep}{3pt}
\renewcommand{\arraystretch}{1.16}
\begin{tabular*}{\linewidth}{@{\extracolsep{\fill}}>{\raggedright\arraybackslash}p{0.31\linewidth}cc>{\centering\arraybackslash}p{0.14\linewidth}>{\centering\arraybackslash}p{0.15\linewidth}>{\centering\arraybackslash}p{0.16\linewidth}@{}}
\toprule
\textbf{Work} & \multicolumn{2}{c}{\textbf{Improvement Scope}} & \textbf{\shortstack{Open-Ended\\Goals}} & \textbf{\shortstack{Benchmark\\Feedback}} & \textbf{\shortstack{Self as\\Improver}}\\
\cmidrule(lr){2-3}
 & Model & Harness & & & \\
\midrule
Meta-Harness \citep{lee2026metaharness} & $\times$ & $\checkmark$ & $\times$ & $\checkmark$ & $\times$\\
RSIBench-Data \citep{meng2026rsibenchdata} & $\checkmark$ & $\times$ & $\times$ & $\checkmark$ & $\times$\\
Co-Harness \citep{chen2026coharness} & $\checkmark$ & $\checkmark$ & $\times$ & $\checkmark$ & $\times$\\
RSIGym \citep{meng2026rsigym} & $\checkmark$ & $\checkmark$ & $\times$ & $\checkmark$ & $\times$\\
Harness$^2$ \citep{xu2026harness2} & $\times$ & $\checkmark$ & $\times$ & $\times$ & $\times$\\
\midrule
\textbf{RSI-Env (ours)} & $\boldsymbol{\checkmark}$ & $\boldsymbol{\checkmark}$ & $\boldsymbol{\checkmark}$ & $\boldsymbol{\times}$ & $\boldsymbol{\checkmark}$\\
\bottomrule
\end{tabular*}
\end{table}

%% file: section/abstract.tex
\begin{abstract}
Recursive self-improvement (RSI) envisions agent systems that autonomously improve their own models and harnesses, with each updated system continuing to improve itself.
Existing approaches typically restrict optimization objectives, editable components, or evaluation procedures, limiting open-ended exploration and continuity across generations.
We introduce \textbf{\rsienv{}}, a persistent environment for coordinated model--harness improvement, supporting autonomous training, harness modification, and retention of research state.
An independent Host Controller manages services, evaluation, and versioning.
Under \emph{Self}, each accepted model--harness version takes over the next improvement round.
We define three settings that vary goal specificity and access to benchmark feedback: \emph{Open-General}, \emph{Capability-Directed}, and \emph{Benchmark-Guided}.
Across these three settings, we compare Self with two fixed external Teachers on seven tasks, starting from Qwen3.8-27B with a minimal harness.
Without benchmark feedback, Open-General Self raises GPQA Diamond accuracy from 65.2\% to 73.2\% and AIME 2026 accuracy from 53.3\% to 56.7\%.
All three Capability-Directed Self runs achieve target-aligned gains.
Knowledge Self demonstrates continued improvement across system updates, with SimpleQA accuracy rising from 82.0\% to 85.0\% and then 88.0\%.
Benchmark guidance and Teacher benefits vary by task.
These results demonstrate \rsienv{}'s support for open-ended exploration and sustained recursive improvement within the studied runs, while highlighting challenges in preserving gains across capabilities.
\end{abstract}

%% file: section/introduction.tex
\section{Introduction}
\begingroup
\widowpenalty=10000

Large language models (LLMs) increasingly support agents that perform complex tasks through reasoning, tool use, and environmental feedback \citep{yao2023react,schick2023toolformer,dong2026longhorizon}.
Their capabilities depend on both the underlying model and the surrounding harness, including prompts, context management, tool policies, and execution logic.
Improving these systems therefore involves decisions about model training, harness design, and the coordination between them \citep{chen2026coharness}.
Such decisions remain largely human-directed.
Recursive self-improvement (RSI) envisions agents taking on this process themselves: autonomously improving their own models and harnesses, with each updated system continuing to improve itself.
This raises a central question: \emph{how can we support agents in organizing and sustaining open-ended improvement of their own systems?}

Recent work has made progress by enabling agents to construct training data \citep{meng2026rsibenchdata}, refine training algorithms \citep{chi2026ai4ai}, modify harnesses \citep{lee2026metaharness,qu2026she,zhang2025dgm,zhang2025ace}, and jointly optimize model weights and agent workflows \citep{chen2026coharness,mao2026safeevolve}.
These studies demonstrate that agents can perform substantial improvement work.
However, the objective, editable components, update procedure, or evaluation feedback is typically specified in advance \citep{mahmoud2026rsibench,meng2026rsibenchdata,chi2026ai4ai,lee2026metaharness}.
This makes it difficult to study how agents formulate an improvement agenda, coordinate changes across system components, and carry the resulting research forward through updated versions of themselves.
Investigating this broader process requires infrastructure that connects autonomous exploration, system updates, and continued research across generations.

Two requirements follow.
First, the environment must support system-level exploration and recursive continuity.
Agents need access to model training and harness modification within a shared workspace, together with persistent records and versioned systems that allow research to continue after an update.
Second, evaluation must reveal how much of the research process agents can organize autonomously.
When a benchmark and its feedback are provided, an agent can pursue an externally defined objective.
With less guidance, it must also identify a worthwhile direction, formulate a concrete problem, and develop its own evidence for assessing candidate changes.
These requirements motivate a common environment in which research autonomy can be varied while recursive continuation remains supported.

We introduce \textbf{\rsienv{}}, a persistent environment for open-ended, system-level self-improvement.
Agents can inspect infrastructure documentation, modify harness code, construct training data, choose training methods, and test candidate systems.
Model and harness changes share the same exploration space, allowing agents to coordinate interventions without following a prescribed update sequence.
Experimental records and workspace state persist across iterations, and accepted model--harness versions are retained.
An independent Host Controller manages services, validity checks, versioning, and capability evaluation outside the editable environment.
Acceptance establishes that an update is valid and executable; capability changes are measured independently.
Under \emph{Self}, the evolving target system itself serves as the Improver, with each accepted model--harness version taking over the next round.
Under \emph{Teacher}, an external Improver updates the target system while its own model and harness remain fixed.
The two Teacher configurations use Codex with GPT-5.6 Sol or GPT-6 Astra.

Three settings vary research autonomy: \emph{Open-General} leaves the improvement direction open, \emph{Capability-Directed} specifies a broad reasoning, knowledge, or agentic goal, and \emph{Benchmark-Guided} supplies a benchmark target and evaluation feedback.
The first two provide neither a named benchmark nor benchmark feedback.
Independent capability evaluations and recorded trajectories assess autonomous improvement, the role of guidance, and differences between Self and external Teachers.

Starting from Qwen3.8-27B with a minimal harness, we compare these configurations on seven tasks spanning the three settings.
Knowledge Self demonstrates continued improvement across system updates, with SimpleQA accuracy rising from 82.0\% to 85.0\% and then 88.0\%.
Without benchmark feedback, Open-General Self raises GPQA Diamond accuracy from 65.2\% to 73.2\% and AIME 2026 accuracy from 53.3\% to 56.7\%.
All three Capability-Directed Self runs achieve target-aligned gains.
Benchmark guidance yields higher Self scores on AIME 2026 and Terminal-Bench 2.1 than the corresponding Capability-Directed settings, but a lower score on SimpleQA.
External Teachers do not consistently outperform Self, and gains on individual benchmarks coexist with regressions elsewhere.

\Needspace{5\baselineskip}
Our contributions are threefold:

\begin{itemize}[leftmargin=*,itemsep=2pt,topsep=2pt]
    \item \textbf{An environment for open-ended RSI.} \rsienv{} supports coordinated model--harness improvement, persistent research state, and continued research by updated systems.
    \item \textbf{A framework for studying research autonomy.} Three settings vary goal specificity and benchmark feedback within a shared infrastructure.
    \item \textbf{An empirical demonstration of sustained RSI.} Trajectories and evaluations show recursive continuation and selective capability gains, alongside challenges in preserving performance across capabilities.
\end{itemize}
\endgroup

%% file: section/related_work.tex
\section{Related Work}

\subsection{Self-Improving Agent Systems}

Agent self-improvement involves model training, harness modification, or their joint optimization. STaR \citep{zelikman2022star} fine-tunes models on self-generated rationales that yield correct answers, and SEAL \citep{zweiger2025seal} generates fine-tuning data and specifies the training settings used to update model weights. RSIBench-Data \citep{meng2026rsibenchdata} evaluates agents that design training-data strategies within a fixed training pipeline. Meta-Harness \citep{lee2026metaharness} and Agentic Harness Engineering \citep{lin2026ahe} modify harness code based on evaluation results and execution traces. Harness$^2$ \citep{xu2026harness2} modifies harnesses for unseen tasks without benchmark feedback and retains prior harnesses and improvement notes for reuse, while model weights remain fixed. Co-Harness \citep{chen2026coharness} alternates between harness modification and model training on trajectories generated by the modified harness. RSI-Env supports model training and harness modification in a shared environment, preserves research records, and uses accepted model--harness systems to conduct subsequent improvement rounds.

\suppressfloats[t]
\input{section/framework_figure}

\subsection{Evaluating Autonomy in Self-Improvement}

Existing benchmarks evaluate agents on predefined research and engineering tasks. RSI Bench \citep{mahmoud2026rsibench} provides AI R\&D tasks with compute budgets and verifiers; PostTrainBench \citep{rank2026posttrainbench} specifies a target model and benchmark; Agent$^2$ RL-Bench \citep{chen2026agent2rlbench} tests agents' ability to design and execute reinforcement-learning training pipelines. RSIGym \citep{meng2026rsigym} provides training, inference, and evaluation services for model training and harness modification, but its reported experiments do not test updated systems taking over subsequent research rounds. RSI-Forge \citep{wang2026rsiforge} constructs executable research tasks from papers and evaluates three successive sessions that inherit solution code and notes, with fixed model weights and a common Claude Code harness. RSI-Env evaluates continued research by updated model--harness systems while varying goal specificity and access to benchmark feedback, including open-ended goals without benchmark feedback. Table~\ref{tab:related-comparison} compares the Open-General and Capability-Directed RSI-Env settings with representative prior work.

%% file: section/framework_figure.tex
\begin{figure}[t]
    \centering
    \includegraphics[width=\linewidth]{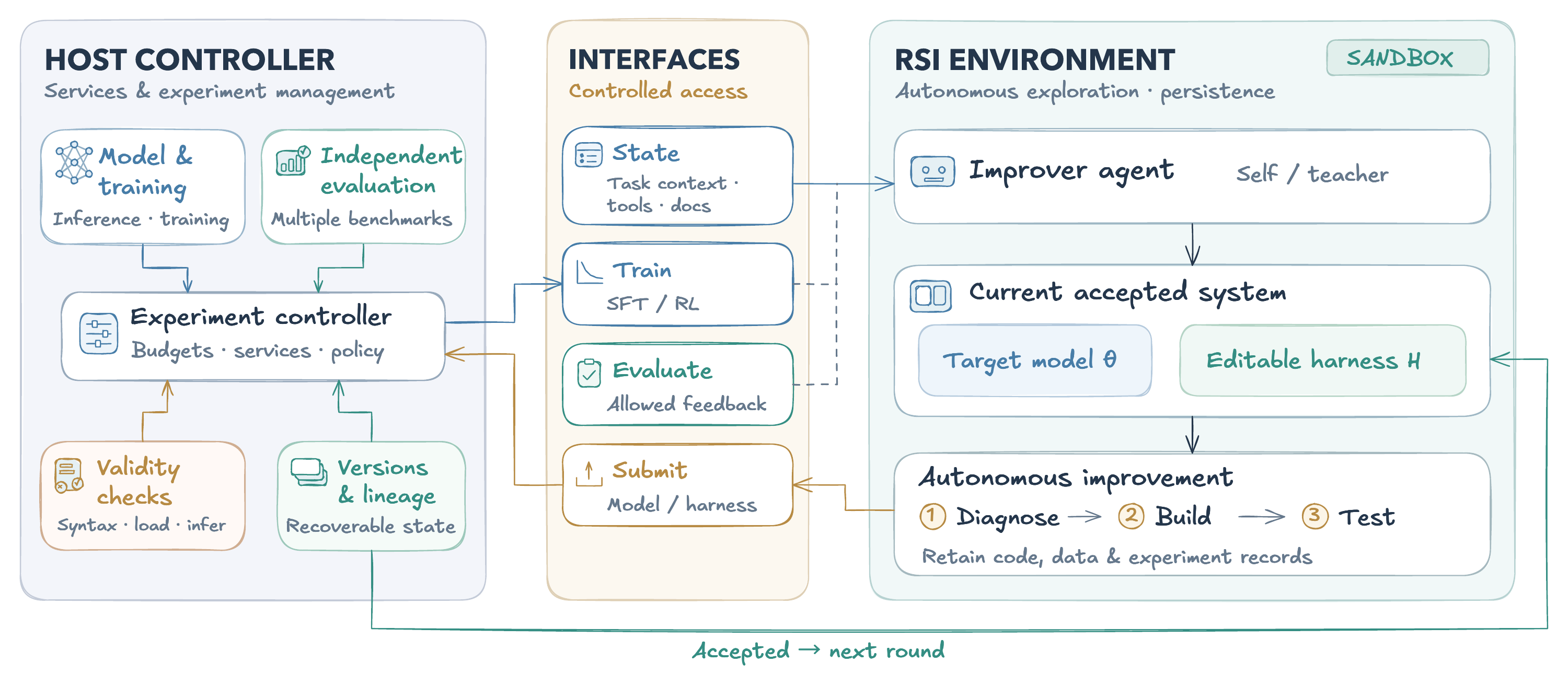}
\caption{RSI-Env architecture. The RSI Environment supports autonomous open-ended exploration in a sandbox and retains working results. The Host Controller provides services through controlled interfaces and maintains private evaluation data and version management.}
\label{fig:rsi-env-framework}
\end{figure}

%% file: section/method.tex
\section{Method}

\subsection{Problem Setup}

We formulate RSI as the iterative improvement of an executable agent system, $\mathcal{S}_t=(\theta_t,H_t,\mathcal{P}_t;\mathcal{R})$. Here, $\theta_t$ denotes the target model parameters and $H_t$ the current executable harness, including prompts, tool policies, context organization, memory, and runtime logic. The persistent state $\mathcal{P}_t$ contains experimental code, records, and intermediate artifacts retained across iterations. The resources $\mathcal{R}$ include external libraries, inference and training services, evaluation services, network access, and computing infrastructure.

Given an improvement goal $G$, an improvement agent (the \emph{Improver}) examines the current system and determines whether to modify the model, the harness, or both. Let $E_t$ denote the subset of $\{\theta_t,H_t\}$ selected at iteration $t$, and let $\Delta_t$ denote the proposed modification, such as a training procedure or a harness code change. A single update is expressed as
\begin{equation}
    u_t=(E_t,\Delta_t),\quad E_t\subseteq\{\theta_t,H_t\},
    \qquad \mathcal{S}_{t+1}\sim\mathcal{T}(\cdot\mid\mathcal{S}_t,u_t),
    \label{eq:rsi-transition}
\end{equation}
where $\mathcal{T}$ denotes the transition distribution over successor systems induced by executing the update, allowing for randomness in training or tool execution. In each iteration, the Improver inspects the current system and retained records, conducts experiments in the sandbox, and may submit a candidate model--harness pair. The Host Controller checks whether the candidate is valid and executable before registering it as the successor; failed attempts and diagnostic results are also retained in $\mathcal{P}_{t+1}$. Candidates are registered based on validity and executability, while capability changes are assessed independently.

In the \emph{Self} configuration, the evolving target model--harness pair also serves as the Improver, so each registered successor generates the next update. In a \emph{Teacher} configuration, a fixed external Improver updates the evolving target system.

\subsection{RSI-Env}

RSI-Env is organized into an \textbf{RSI Environment} for autonomous exploration and a \textbf{Host Controller} for service provision and experiment management. Communication between these components is mediated by \textbf{controlled interfaces}. Candidate modifications are developed in the RSI Environment, where experimental results are retained across iterations. In the Host Controller, model and training services, independent evaluation, validity checks, and version and lineage management are coordinated by the Experiment Controller. The overall architecture is shown in \rsifigref{fig:rsi-env-framework}.

\paragraph{RSI Environment.}
Autonomous exploration is carried out by an improvement agent (\emph{Improver}) within a sandbox, where modifications to the model or harness can be developed and tested. State is retained across iterations, while experimental changes are kept separate from the accepted system. Consequently, evidence from unsuccessful iterations can be reused in subsequent exploration, and work can be resumed from an accepted version without discarding accumulated records.

\paragraph{Host Controller.}
Training services, independent evaluation, validity checks, and version and lineage management are coordinated outside the editable environment. Evaluation criteria and version records are protected from modification by the Improver. Execution diagnostics, such as service errors and training outputs, may be returned during exploration; task-specific evaluation feedback is exposed only when permitted by the task configuration. Independent capability evaluation remains controller-side and is not used as the acceptance criterion for intermediate updates. Retained lineage and experiment records link observed outcomes to the corresponding updates and execution stages.

\paragraph{Controlled interfaces.}
Access to controller-managed services is mediated by controlled interfaces, through which authorized resources can be used while the choice of improvement approach is left to the Improver. The experimenter can configure which services and evaluation feedback are exposed for a task without changing the underlying environment.

\begin{figure}[!t]
\centering
\includegraphics[width=\linewidth]{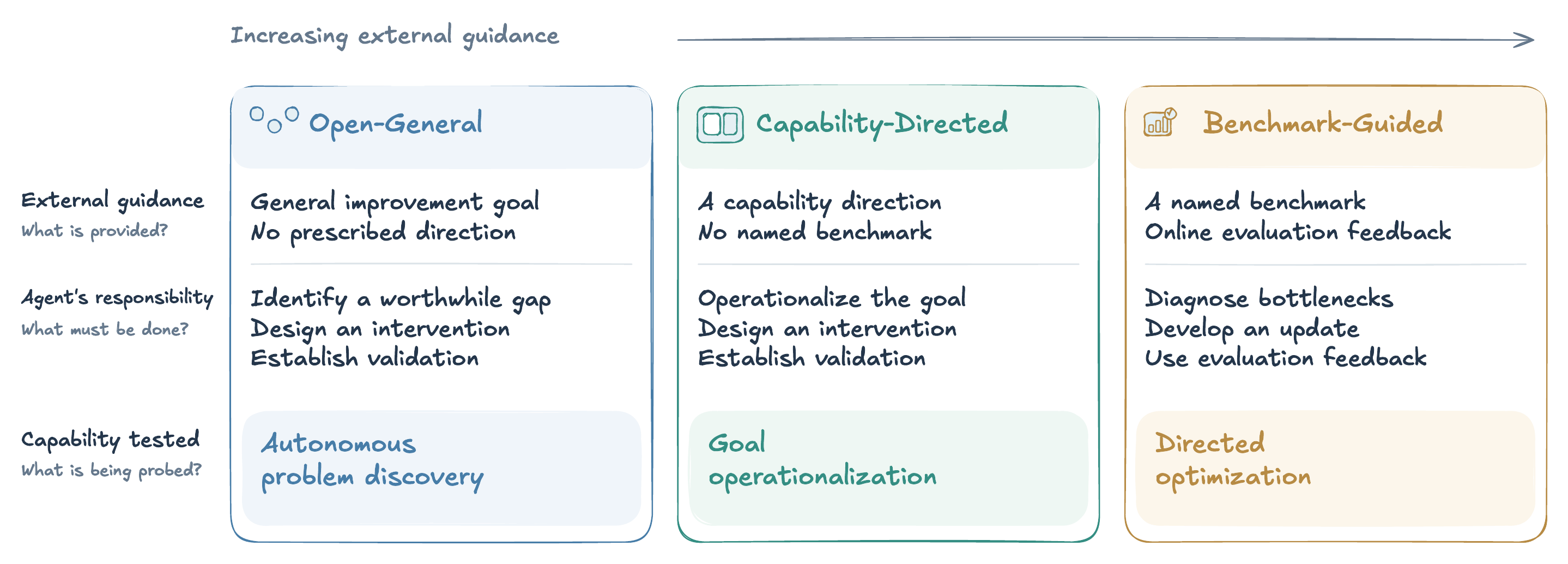}
\caption{Three task settings for assessing progress toward RSI, varying goal specificity and the availability of evaluation feedback. Increasing external guidance reduces the problem-formulation responsibilities assigned to the agent.}
\label{fig:task-contracts}
\end{figure}

\subsection{Task Design for Assessing Progress toward RSI}
\label{sec:task-design}

Each RSI-Env task specifies an improvement goal, the controller-managed services and task-specific feedback exposed to the Improver, and an independent final evaluation. These elements can be varied while holding the underlying environment and core resource pool fixed. The three settings used in this work instantiate different points along this spectrum and examine \emph{how much of the improvement process can be organized autonomously}.

The settings progressively transfer problem-formulation responsibility to the Improver (\rsifigref{fig:task-contracts}). \emph{Benchmark-Guided} specifies a named target and provides corresponding evaluation feedback. \emph{Capability-Directed} provides only a broad capability goal, without a named benchmark or benchmark feedback, requiring the Improver to formulate a concrete problem and establish local evidence of progress. \emph{Open-General} also leaves the improvement direction unspecified. Basic execution diagnostics remain available in all settings, while final capability gains are measured independently and are not used to accept intermediate updates.

Across the three settings, we use the same minimal initial system, consisting of a language model and a simple harness that implements only a basic agent loop. Starting from the same foundation, the Improver must develop an update strategy for a named target, formulate and validate a problem within a capability area, or additionally select the improvement direction itself. Independent capability measurements and recorded trajectories are used to examine both the resulting gains and how the system develops and carries forward its improvement process.

%% file: section/experiments.tex
\section{Experiments}
\label{sec:experiments}

\subsection{Experimental Setup}

\paragraph{Task settings.}
The three settings in Section~\ref{sec:task-design} vary the specificity of the improvement goal and the availability of evaluation feedback. \textbf{Open-General} leaves the improvement direction open. \textbf{Capability-Directed} specifies reasoning, knowledge, or agentic capability as the improvement goal. These goals capture three important aspects of agent performance: solving problems through reasoning, acquiring and applying factual knowledge, and using tools to act in an environment \textcolor{blue}{\citep{liu2023agentbench,dong2026longhorizon}}. \textbf{Benchmark-Guided} provides concrete targets corresponding to these capabilities: AIME 2026 for mathematical reasoning \textcolor{blue}{\citep{maa2025aime2026}}, SimpleQA for factual knowledge \textcolor{blue}{\citep{wei2024simpleqa}}, and Terminal-Bench 2.1 for tool use and task execution \textcolor{blue}{\citep{merrill2026terminalbench}}. Every experiment is allocated the same budget: ten rounds of three hours each, totaling thirty hours. The Improver may finish a round early; reported runtime is actual time used, not the allocated budget. Detailed experimental settings and task definitions are provided in Appendices~\ref{app:settings-tasks} and~\ref{app:task-definitions}.

\paragraph{Initial agent and Improver.}
Qwen3.8-27B \textcolor{blue}{\citep{qwen2026qwen38}} is used as the initial target model and paired with a deliberately minimal harness, which implements a basic agent loop and is detailed alongside the editable components in Appendix~\ref{app:initial-system}. Three Improver configurations are used in the experiments: \emph{Self} and two external \emph{Teachers}. Under \emph{Self}, this initial model-harness pair also serves as the Improver. After each accepted update, the resulting pair takes over the next improvement iteration, carrying the evolving model and harness forward until the run ends. Under \emph{Teacher}, an external Improver modifies the target system while its own model and harness remain fixed throughout the run. The two Teacher configurations use Codex with either GPT-5.6 Sol \textcolor{blue}{\citep{openai2026gpt56}} or GPT-6 Astra, both with High reasoning effort.

\paragraph{Training support.}
Model training is supported through the Tinker SDK \textcolor{blue}{\citep{thinkingmachines2025tinker}}, which provides supervised fine-tuning and the operations needed to construct custom reinforcement learning workflows. Implementation details are provided in Appendix~\ref{app:training}.

\paragraph{Evaluation.}
The resulting model-harness pairs are evaluated using Harbor \textcolor{blue}{\citep{terminalbench2025harbor}} on four benchmarks covering complementary capabilities. Each item is evaluated three times; we report avg@3 for GPQA Diamond, AIME 2026, and SimpleQA, and pass@3 for Terminal-Bench 2.1. GPQA Diamond is the 198-question high-quality subset of GPQA and assesses scientific knowledge and reasoning \textcolor{blue}{\citep{rein2024gpqa}}. AIME 2026 comprises 30 competition mathematics problems from the American Invitational Mathematics Examination and assesses mathematical problem solving. Terminal-Bench 2.1 contains 89 tasks in terminal environments and assesses agents' ability to carry out practical computational work. For SimpleQA, a randomly sampled subset of 500 short, fact-seeking questions is used to assess factual accuracy. The evaluation framework and hyperparameter settings are described in Appendix~\ref{app:evaluation}.

\begin{table}[!t]
\centering
\fontsize{8.5}{10.5}\selectfont
\setlength{\tabcolsep}{1.9pt}
\renewcommand{\arraystretch}{1.12}
\setlength{\heavyrulewidth}{0.65pt}
\setlength{\lightrulewidth}{0.35pt}
\setlength{\cmidrulewidth}{0.25pt}
\setlength{\aboverulesep}{1.8pt}
\setlength{\belowrulesep}{1.8pt}
\arrayrulecolor{black!45}
\definecolor{deltaGain}{HTML}{2E8B57}
\definecolor{deltaLoss}{HTML}{C23B3B}
\definecolor{deltaZero}{HTML}{777777}
\definecolor{targetBlue}{HTML}{E7F2FB}
\newcommand{\targetcell}[1]{\cellcolor{targetBlue}#1}
\newcommand{\deltagain}[1]{\textcolor{deltaGain}{#1}}
\newcommand{\deltaloss}[1]{\textcolor{deltaLoss}{#1}}
\newcommand{\deltazero}[1]{\textcolor{deltaZero}{#1}}
\newcommand{\outcomeheader}[1]{{\bfseries\raisebox{-0.5\height}{\shortstack{#1}}}}
\caption{Self and Teacher results across Open-General and Capability-Directed settings. Light blue cells mark target-aligned benchmarks.}
\label{tab:outcomes}
\begin{tabular}{@{}l@{\hspace{6pt}}l@{\hspace{6pt}}*{4}{cc}@{\hspace{5pt}}ccc@{}}
\toprule
 & & \multicolumn{2}{c}{\outcomeheader{GPQA\\Diamond}} & \multicolumn{2}{c}{\outcomeheader{AIME\\2026}} & \multicolumn{2}{c}{\outcomeheader{Terminal-Bench\\2.1}} & \multicolumn{2}{c}{\outcomeheader{SimpleQA}} & \multicolumn{3}{c}{\outcomeheader{Improvement process}} \\
\cmidrule(lr){3-4}\cmidrule(lr){5-6}\cmidrule(lr){7-8}\cmidrule(lr){9-10}\cmidrule(lr){11-13}
\textbf{Task} & \textbf{Improver} & \shortstack{avg@3\\(\%)} & $\Delta$ & \shortstack{avg@3\\(\%)} & $\Delta$ & \shortstack{pass@3\\(\%)} & $\Delta$ & \shortstack{avg@3\\(\%)} & $\Delta$ & \shortstack{Final version\\(H/M)} & \shortstack{Tool calls\\(total)} & \shortstack{Total time\\(h)} \\
\midrule
\textbf{base} & seed & 65.2 & -- & 53.3 & -- & 12.4 (2.2) & -- & 82.0 & -- & H0/M0 & n/a & n/a \\
\midrule
\multicolumn{13}{@{}l}{\textsc{Open-General}} \\
\textbf{General} & Self & 73.2 & \deltagain{$\uparrow\!8.0$} & 56.7 & \deltagain{$\uparrow\!3.3$} & 9.0 (2.2) & \deltaloss{$\downarrow\!3.4$} & 77.2 & \deltaloss{$\downarrow\!4.8$} & H1/M0 & 454 & 28.95 \\
 & GPT-5.6 Sol & 60.6 & \deltaloss{$\downarrow\!4.6$} & 36.7 & \deltaloss{$\downarrow\!16.6$} & 9.0 (1.1) & \deltaloss{$\downarrow\!3.4$} & 81.0 & \deltaloss{$\downarrow\!1.0$} & H10/M0 & 205 & 1.27 \\
 & GPT-6 Astra & 70.7 & \deltagain{$\uparrow\!5.6$} & 60.0 & \deltagain{$\uparrow\!6.7$} & 11.2 (3.4) & \deltaloss{$\downarrow\!1.1$} & 79.8 & \deltaloss{$\downarrow\!2.2$} & H4/M0 & 902 & 6.97 \\
\midrule
\multicolumn{13}{@{}l}{\textsc{Capability-Directed}} \\
\textbf{Reasoning} & Self & \targetcell{69.2} & \targetcell{\deltagain{$\uparrow\!4.0$}} & \targetcell{53.3} & \targetcell{\deltazero{$0.0$}} & 6.7 (1.1) & \deltaloss{$\downarrow\!5.7$} & 80.0 & \deltaloss{$\downarrow\!2.0$} & H1/M1 & 161 & 25.57 \\
 & GPT-5.6 Sol & \targetcell{58.6} & \targetcell{\deltaloss{$\downarrow\!6.6$}} & \targetcell{56.7} & \targetcell{\deltagain{$\uparrow\!3.3$}} & 6.7 (1.1) & \deltaloss{$\downarrow\!5.7$} & 81.6 & \deltaloss{$\downarrow\!0.4$} & H10/M0 & 279 & 1.19 \\
 & GPT-6 Astra & \targetcell{68.7} & \targetcell{\deltagain{$\uparrow\!3.5$}} & \targetcell{46.7} & \targetcell{\deltaloss{$\downarrow\!6.6$}} & 10.1 (1.1) & \deltaloss{$\downarrow\!2.3$} & 78.6 & \deltaloss{$\downarrow\!3.4$} & H3/M0 & 705 & 3.91 \\
\midrule[0.25pt]
\textbf{Agentic} & Self & 70.2 & \deltagain{$\uparrow\!5.0$} & 70.0 & \deltagain{$\uparrow\!16.7$} & \targetcell{14.6 (2.2)} & \targetcell{\deltagain{$\uparrow\!2.2$}} & 80.0 & \deltaloss{$\downarrow\!2.0$} & H7/M0 & 269 & 24.90 \\
 & GPT-5.6 Sol & 54.5 & \deltaloss{$\downarrow\!10.7$} & 53.3 & \deltazero{$0.0$} & \targetcell{7.9 (0.0)} & \targetcell{\deltaloss{$\downarrow\!4.5$}} & 84.6 & \deltagain{$\uparrow\!2.6$} & H10/M0 & 245 & 1.22 \\
 & GPT-6 Astra & 66.2 & \deltagain{$\uparrow\!1.0$} & 53.3 & \deltazero{$0.0$} & \targetcell{9.0 (1.1)} & \targetcell{\deltaloss{$\downarrow\!3.4$}} & 82.6 & \deltagain{$\uparrow\!0.6$} & H7/M0 & 597 & 4.04 \\
\midrule[0.25pt]
\textbf{Knowledge} & Self & 53.0 & \deltaloss{$\downarrow\!12.2$} & 66.7 & \deltagain{$\uparrow\!13.4$} & 5.6 (0.0) & \deltaloss{$\downarrow\!6.8$} & \targetcell{88.0} & \targetcell{\deltagain{$\uparrow\!6.0$}} & H1/M1 & 192 & 27.79 \\
 & GPT-5.6 Sol & 56.6 & \deltaloss{$\downarrow\!8.6$} & 53.3 & \deltazero{$0.0$} & 13.5 (1.1) & \deltagain{$\uparrow\!1.1$} & \targetcell{86.6} & \targetcell{\deltagain{$\uparrow\!4.6$}} & H8/M0 & 266 & 1.24 \\
 & GPT-6 Astra & 67.2 & \deltagain{$\uparrow\!2.0$} & 50.0 & \deltaloss{$\downarrow\!3.3$} & 6.7 (2.2) & \deltaloss{$\downarrow\!5.7$} & \targetcell{82.0} & \targetcell{\deltazero{$0.0$}} & H6/M0 & 730 & 4.88 \\
\bottomrule
\end{tabular}
\end{table}

\subsection{Autonomous Improvement under Open-Ended Goals}
\label{sec:autonomous-outcomes}

\textbf{Self achieves autonomous capability gains without benchmark feedback} (Table~\ref{tab:outcomes}). Under Open-General, Self raises GPQA Diamond accuracy from 65.2\% to 73.2\% and AIME 2026 accuracy from 53.3\% to 56.7\% through a harness update with unchanged model weights (H1/M0), then continues local evaluation and failure analysis. \textbf{All three Capability-Directed Self runs achieve target-aligned gains:} Reasoning Self reaches 69.2\% on GPQA Diamond with AIME 2026 unchanged at 53.3\%; Agentic Self raises Terminal-Bench 2.1 pass@3 from 12.4\% to 14.6\%; and Knowledge Self raises SimpleQA accuracy from 82.0\% to 88.0\%. \textbf{Gains coexist with regressions on other benchmarks:} Open-General Self falls to 77.2\% on SimpleQA, and Knowledge Self falls to 53.0\% on GPQA Diamond.

\subsection{Goal Specification and Benchmark Feedback}
\label{sec:guidance-outcomes}

\begin{figure}[!htb]
\centering
\includegraphics[width=\linewidth]{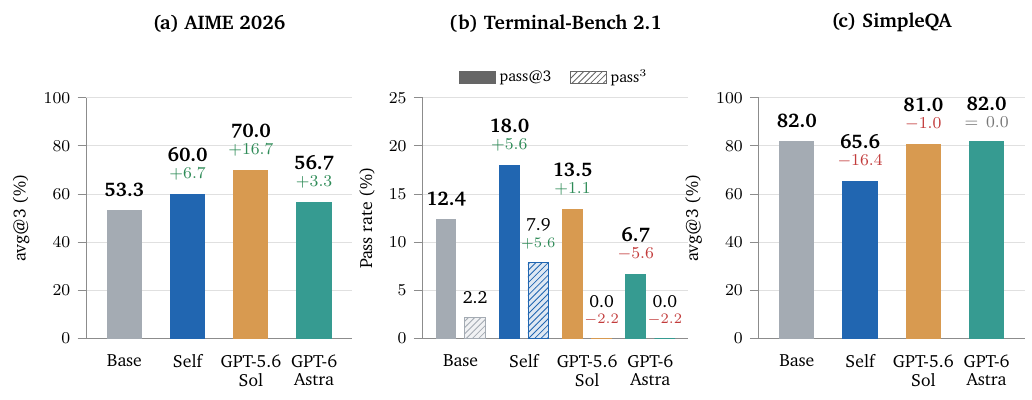}
\caption{Benchmark-Guided results. Gray bars show the initial system; colored labels show score changes from that system.}
\label{fig:benchmark-guided-outcomes}
\end{figure}

\textbf{Benchmark-Guided Self outperforms its Capability-Directed counterparts on AIME 2026 and Terminal-Bench 2.1} (Table~\ref{tab:outcomes}; Figure~\ref{fig:benchmark-guided-outcomes}). Self's Terminal-Bench 2.1 pass@3 rises from 9.0\% under Open-General to 14.6\% under the agentic goal and 18.0\% under Benchmark-Guided; tasks passed in all three trials rise from 2.2\% under the agentic goal to 7.9\% under Benchmark-Guided. On AIME 2026, Benchmark-Guided Self reaches 60.0\%, compared with 53.3\% under the reasoning goal and 56.7\% under Open-General. \textbf{SimpleQA is the exception in these runs:} Self scores 77.2\% under Open-General, 88.0\% under the knowledge goal, and 65.6\% under Benchmark-Guided.

\subsection{Self and External Teachers}
\label{sec:teacher-outcomes}
\label{sec:resource-use}

\textbf{Counterintuitively, stronger models are not consistently better improvers} (Table~\ref{tab:outcomes}). Under Capability-Directed, Self reaches 69.2\% on GPQA Diamond for reasoning, 14.6\% on Terminal-Bench 2.1 pass@3 for agentic capability, and 88.0\% on SimpleQA for knowledge, exceeding both Teachers on these measures. Teacher advantages include Astra's 60.0\% on AIME 2026 under Open-General versus Self's 56.7\%, and Sol's 70.0\% under Benchmark-Guided versus Self's 60.0\%; both Teachers also outperform Benchmark-Guided Self on SimpleQA. Conversely, Benchmark-Guided Self leads Terminal-Bench 2.1 pass@3 at 18.0\%, versus 13.5\% for Sol and 6.7\% for Astra.

\textbf{Models differ substantially in how they use the improvement budget.} Across Open-General and Capability-Directed, Self uses 24.90--28.95 hours, Sol 1.19--1.27 hours, and Astra 3.91--6.97 hours under the same 30-hour allocation. Within these runs, only Reasoning and Knowledge Self accept model updates; Teacher endpoints retain M0 despite multiple harness updates. \textbf{Differences in time use and model-update behavior may partly explain why the stronger Teachers do not consistently outperform Self.}

\Needspace{10\baselineskip}
\subsection{Case Study: Sustained Recursive Improvement}
\label{sec:knowledge-case}

\definecolor{caseBlue}{HTML}{2166B1}
\definecolor{caseTeal}{HTML}{369B91}
\definecolor{caseOrange}{HTML}{D89A50}
\definecolor{caseGray}{HTML}{A4ABB3}
\definecolor{caseGreen}{HTML}{2E8B57}
\definecolor{caseRed}{HTML}{C23B3B}
\begin{figure}[!htb]
\centering
\begin{minipage}[t]{0.40\linewidth}
\centering
\begin{minipage}[c][165pt][c]{\linewidth}
\centering
\includegraphics[width=\linewidth]{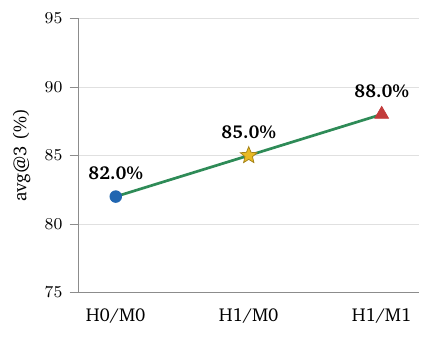}
\end{minipage}\par\smallskip
\textbf{(a)}
\end{minipage}\hfill
\begin{minipage}[t]{0.55\linewidth}
\centering
\begin{minipage}[c][165pt][c]{\linewidth}
\centering
\fontsize{8}{9.5}\selectfont
\textbf{Data quality}\par\smallskip
\begin{tikzpicture}[x=0.11cm,y=0.45cm]
\fill[caseGreen] (0,0) rectangle (43,0.55);
\fill[caseOrange] (43,0) rectangle (46,0.55);
\fill[caseRed] (46,0) rectangle (48,0.55);
\node[text=white,font=\scriptsize] at (21.5,0.275) {43};
\node[font=\scriptsize] at (44.5,0.275) {3};
\node[font=\scriptsize] at (47,0.275) {2};
\end{tikzpicture}\par
{\scriptsize\textcolor{caseGreen}{Acceptable: 43}\quad
\textcolor{caseOrange}{Minor issues: 3}\quad
\textcolor{caseRed}{Defective: 2}}\par\smallskip
\textbf{Data-type distribution}\par\smallskip
\begin{minipage}[c]{0.34\linewidth}
\centering
\begin{tikzpicture}[scale=0.95]
\fill[caseBlue] (0,0) -- (90:1) arc (90:210:1) -- cycle;
\fill[caseTeal] (0,0) -- (210:1) arc (210:285:1) -- cycle;
\fill[caseBlue!45] (0,0) -- (285:1) arc (285:390:1) -- cycle;
\fill[caseOrange] (0,0) -- (390:1) arc (390:420:1) -- cycle;
\fill[caseGray] (0,0) -- (420:1) arc (420:450:1) -- cycle;
\fill[white] (0,0) circle (0.58);
\node[font=\small\bfseries] at (0,0) {48};
\end{tikzpicture}
\end{minipage}\hfill
\begin{minipage}[c]{0.65\linewidth}
\fontsize{6.8}{8.2}\selectfont
\setlength{\tabcolsep}{2pt}
\begin{tabular}{@{}clr@{}}
\textcolor{caseBlue}{\rule{5pt}{5pt}} & Definitions, units, quantities & 16\\
\textcolor{caseTeal}{\rule{5pt}{5pt}} & Constraints and formats & 10\\
\textcolor{caseBlue!45}{\rule{5pt}{5pt}} & Code, planning, general ability & 14\\
\textcolor{caseOrange}{\rule{5pt}{5pt}} & Calibration & 4\\
\textcolor{caseGray}{\rule{5pt}{5pt}} & Numerical discipline & 4\\
\end{tabular}
\end{minipage}\par\smallskip
\setlength{\tabcolsep}{4pt}
\begin{tabular}{@{}cccc@{}}
\toprule
\textbf{Method} & \textbf{Examples} & \textbf{Epochs} & \textbf{Training loss}\\
\midrule
SFT & 48 & 2 & $5.9874 \rightarrow 4.5796$\\
\bottomrule
\end{tabular}
\end{minipage}\par\smallskip
\textbf{(b)}
\end{minipage}
\caption{Knowledge Self: (a) SimpleQA avg@3 across checkpoints; (b) composition and quality audit of the 48 training examples.}
\label{fig:checkpoint-progression}
\end{figure}

\textbf{Knowledge Self achieves successive gains as updated systems continue the improvement process} (Figure~\ref{fig:checkpoint-progression}). Starting from H0/M0 at 82.0\% SimpleQA avg@3, the agent updates its harness to obtain H1/M0 at 85.0\%. The updated H1/M0 system continues research, constructs 48 prompt--completion pairs, and trains M1. The resulting H1/M1 system reaches 88.0\%. \textbf{This sequence demonstrates recursive continuation in RSI-Env, with research carried forward by the updated system and followed by another capability gain.}

%% file: section/conclusion.tex
\section{Conclusion}

RSI-Env provides a persistent environment for autonomous model training and harness modification, enabling updated systems to continue research with inherited state. Experiments across three settings demonstrate capability gains without benchmark feedback and successive improvement across system versions. Current progress remains uneven: gains on target benchmarks can coincide with regressions elsewhere, and stronger external improvers do not consistently produce better outcomes. Future work could investigate whether targeted human guidance on research goals, experimental design, and budget allocation helps agents pursue more effective improvements while preserving broader capabilities. RSI-Env provides a common infrastructure for studying this collaboration alongside autonomous RSI.

%% file: section/appendix_contents.tex
\pdfbookmark[0]{Appendix Contents}{appendix-contents}
\section*{Appendix Contents}
\begingroup
\small
\setlength{\parindent}{0pt}
\setlength{\parskip}{4pt}
\newcommand{\appendixentry}[2]{\par\noindent\hspace*{#1}\hyperref[#2]{\ref*{#2}\quad\nameref*{#2}\nobreak\dotfill\nobreak\pageref*{#2}}\par}
{\bfseries\appendixentry{0pt}{app:settings-tasks}}
\appendixentry{12pt}{app:initial-system}
\appendixentry{12pt}{app:improver-budgets}
\appendixentry{12pt}{app:hardware}
\appendixentry{12pt}{app:training}
\appendixentry{12pt}{app:evaluation}
\medskip
{\bfseries\appendixentry{0pt}{app:task-definitions}}
\appendixentry{12pt}{app:general-prompt}
\appendixentry{12pt}{app:capability-prompts}
\appendixentry{26pt}{app:reasoning-prompt}
\appendixentry{26pt}{app:knowledge-prompt}
\appendixentry{26pt}{app:agentic-prompt}
\appendixentry{12pt}{app:benchmark-prompts}
\appendixentry{26pt}{app:terminalbench-prompt}
\appendixentry{26pt}{app:simpleqa-prompt}
\appendixentry{26pt}{app:aime-prompt}
\endgroup
\clearpage

%% file: section/appendix_settings.tex
\definecolor{promptGeneral}{HTML}{6961B8}
\definecolor{promptCapability}{HTML}{348A80}
\definecolor{promptBenchmark}{HTML}{477DA8}

\section{Experimental Configuration and Implementation}
\label{app:settings-tasks}

\subsection{Initial System and Editable Components}
\label{app:initial-system}

The initial system pairs Qwen3.8-27B with a minimal harness that calls the model with the task and interaction history, executes one tool action, and appends its result. The two exposed tools are \texttt{bash} for environment interaction and \texttt{compact} for model-generated history summaries. Automatic compaction starts near 90\% context usage and retains the original task. Full tool outputs, workspace files, journals, and experiment records persist across iterations. Harness changes must preserve the inference and execution interfaces. Model training uses controller-managed services; inference backends, credentials, evaluation, and version management remain outside the editable environment.

\subsection{Improver Configuration and Run Budgets}
\label{app:improver-budgets}

Self starts from the target model and minimal harness; accepted model--harness versions conduct subsequent rounds. The fixed Teachers use Codex with GPT-5.6 Sol or GPT-6 Astra, both at High reasoning effort. Each run receives ten three-hour rounds and may finish early. Table~\ref{tab:outcomes} reports actual runtime and cumulative tool calls.

\subsection{Hardware}
\label{app:hardware}

The cluster contains eight nodes, each with eight NVIDIA A800-SXM4-80GB GPUs. Training and inference each use four nodes, totaling 32 GPUs per service.

\subsection{Training Interface}
\label{app:training}

The Tinker SDK is accessed through a Host Controller relay. The \texttt{train-sft} interface accepts JSONL prompt--completion pairs and training settings, then returns losses and a registered model candidate. Sampling, \texttt{forward\_backward}, \texttt{forward\_backward\_custom}, and \texttt{optim\_step} support agent-defined reinforcement learning workflows.

\subsection{Evaluation Framework and Hyperparameters}
\label{app:evaluation}

Table~\ref{tab:main-experiment-settings} lists the budget and sampling settings. The fixed 500-question SimpleQA subset is scored by \texttt{gpt-5.5} \citep{openai2026gpt55}.

\begin{table}[H]
\centering
\caption{Improvement budget and evaluation settings.}
\label{tab:main-experiment-settings}
\small
\setlength{\tabcolsep}{10pt}
\renewcommand{\arraystretch}{1.15}
\begin{tabular}{@{}ll@{}}
\toprule
\textbf{Parameter} & \textbf{Value} \\
\midrule
\multicolumn{2}{@{}l}{\textit{Default improvement budget}} \\
Allocated iterations per run & 10 \\
Maximum duration per iteration & 3 hours (early completion allowed) \\
Total allocated time per run & 30 hours \\
\midrule
\multicolumn{2}{@{}l}{\textit{Evaluation settings}} \\
Trials per item & 3 \\
GPQA Diamond, AIME 2026, SimpleQA metric & avg@3 \\
Terminal-Bench 2.1 metric & pass@3 \\
Context window & 262,144 tokens \\
Automatic context compaction threshold & 235,929 tokens \\
Maximum output per model response & 81,920 tokens \\
Temperature & 1.0 \\
Top-$p$ & 0.95 \\
Top-$k$ & 20 \\
SimpleQA subset & 500 randomly sampled, fixed questions \\
SimpleQA judge & \texttt{gpt-5.5} \\
\bottomrule
\end{tabular}
\end{table}

\paragraph{Metrics.}
GPQA Diamond, AIME 2026, and SimpleQA report avg@3, the average accuracy over three trials. Terminal-Bench 2.1 reports pass@3, the proportion of tasks passed at least once in three trials.

%% file: section/appendix_prompts.tex
\clearpage
\section{Task Prompts}
\label{app:task-definitions}

The seven tasks span the three settings in Section~\ref{sec:task-design}. The Open-General template consolidates the shared instructions; the subsequent boxes specify the blocks replaced for each task.

\tcbset{rsiPrompt/.style={enhanced,breakable,boxrule=0.65pt,arc=1.5mm,left=3mm,right=3mm,top=2mm,bottom=2mm,pad at break*=2mm,fonttitle=\bfseries\fontsize{9}{11}\selectfont,fontupper=\ttfamily\fontsize{8.5}{10.5}\selectfont\raggedright,before upper={\setlength{\parindent}{0pt}\setlength{\parskip}{4pt}},before skip=7pt,after skip=7pt}}

\subsection{Open-General}
\label{app:general-prompt}

\begin{tcolorbox}[rsiPrompt,colback=promptGeneral!4!white,colframe=promptGeneral,colbacktitle=promptGeneral,coltitle=white,title={Open-General Prompt Template},title after break={Open-General Prompt Template (continued)},fontupper=\ttfamily\fontsize{8.5}{10}\selectfont\raggedright,before upper={\setlength{\parindent}{0pt}\setlength{\parskip}{2pt}}]
\textbf{Objective}

Improve the capabilities of the live agent, the model--harness pair, while avoiding regressions in its existing capabilities. Select the improvement direction and establish legitimate local evidence of progress.

\textbf{System updates}

You may train the model, modify the harness, or do both. A controller-accepted replacement ends the current generation. Load-and-sample checks establish executability, not capability improvement. Preserve durable evidence before submitting a replacement.

\textbf{Persistent research state}

At the start of each iteration, inspect the active system, lineage, workspace, journal, and earlier experiment records. Continue from retained evidence rather than repeat completed investigation.

\textbf{Local validation}

Use independently constructed tests, synthetic examples, controlled comparisons, and failure analysis to assess candidate changes. Do not claim an improvement that was not measured.

\textbf{Benchmark access}

No benchmark name or benchmark feedback is supplied. Formal evaluation is controller-owned and unavailable during improvement. Do not locate or access protected benchmark prompts, answers, task identifiers, trajectories, evaluator files, caches, or private service paths. Do not use benchmark-derived material as training data or specialize behavior to known evaluation items. If protected material is accidentally exposed, record the incident and exclude it from training and update decisions.

\textbf{Durable records}

Keep the journal current throughout the iteration. Record hypotheses, observations, local tests, candidate comparisons, changed files, training-data provenance, failures, and conclusions before submitting an update.

\textbf{Budget and completion}

The run allows up to ten iterations of three hours each. If a round ends without an accepted update, resume from the journal and workspace when budget remains. If no further justified improvement is identified, record the conclusion and return exactly TASK\_COMPLETE without submitting a replacement; this ends the run.
\end{tcolorbox}

\subsection{Capability-Directed}
\label{app:capability-prompts}

Replace only the \textbf{Objective} block in Appendix~\ref{app:general-prompt} with the corresponding text below. The shared instructions, including the absence of benchmark names and benchmark feedback, remain applicable.

\subsubsection{Reasoning}
\label{app:reasoning-prompt}

\begin{tcolorbox}[rsiPrompt,colback=promptCapability!4!white,colframe=promptCapability,colbacktitle=promptCapability,coltitle=white,title={Reasoning: Objective}]
Improve reliable reasoning on unfamiliar tasks: identify facts and constraints, plan multi-step deductions, detect failures, use tools appropriately, and verify conclusions. Preserve the agent's other capabilities. Longer reasoning or more verbose answers are not objectives.
\end{tcolorbox}

\subsubsection{Knowledge}
\label{app:knowledge-prompt}

\begin{tcolorbox}[rsiPrompt,colback=promptCapability!4!white,colframe=promptCapability,colbacktitle=promptCapability,coltitle=white,title={Knowledge: Objective}]
Improve accurate knowledge use on unfamiliar questions: retrieve and organize information, retain qualifiers and units, distinguish evidence from assumptions, detect conflicts, and calibrate confidence. Preserve the agent's other capabilities.
\end{tcolorbox}

\subsubsection{Agentic}
\label{app:agentic-prompt}

\begin{tcolorbox}[rsiPrompt,colback=promptCapability!4!white,colframe=promptCapability,colbacktitle=promptCapability,coltitle=white,title={Agentic: Objective}]
Improve completion of unfamiliar multi-step shell and tool tasks through planning, workspace inspection, correct execution, failure recovery, state preservation, and postcondition checks. Preserve the agent's other capabilities.
\end{tcolorbox}

\subsection{Benchmark-Guided}
\label{app:benchmark-prompts}

Replace the \textbf{Objective} and \textbf{Benchmark access} blocks in Appendix~\ref{app:general-prompt}. Use the shared access block below and the objective for the target benchmark; all other template blocks remain applicable.

\begin{tcolorbox}[rsiPrompt,colback=promptBenchmark!4!white,colframe=promptBenchmark,colbacktitle=promptBenchmark,coltitle=white,title={Benchmark-Guided: Benchmark access}]
Request feedback on the named target only through the controller-provided benchmark interface and within the current round budget. Do not obtain benchmark prompts, answers, task identifiers, rollouts, or external solutions from other sources. Do not use benchmark material as training data, specialize the harness to known items, or access protected evaluator files and private service paths. Record accidental exposure and exclude it from training and update decisions. Model load-and-sample checks establish executability, not benchmark improvement.
\end{tcolorbox}

\subsubsection{Terminal-Bench 2.1}
\label{app:terminalbench-prompt}

\begin{tcolorbox}[rsiPrompt,colback=promptBenchmark!4!white,colframe=promptBenchmark,colbacktitle=promptBenchmark,coltitle=white,title={Terminal-Bench 2.1: Objective}]
Improve reliable completion of Terminal-Bench 2.1 tasks under the same per-task budget while retaining reasoning, knowledge, coding, and instruction-following capabilities.
\end{tcolorbox}

\subsubsection{SimpleQA}
\label{app:simpleqa-prompt}

\begin{tcolorbox}[rsiPrompt,colback=promptBenchmark!4!white,colframe=promptBenchmark,colbacktitle=promptBenchmark,coltitle=white,title={SimpleQA: Objective}]
Improve factual accuracy on SimpleQA under the same per-task budget while retaining reasoning, planning, tool use, coding, and instruction-following capabilities.
\end{tcolorbox}

\subsubsection{AIME 2026}
\label{app:aime-prompt}

\begin{tcolorbox}[rsiPrompt,colback=promptBenchmark!4!white,colframe=promptBenchmark,colbacktitle=promptBenchmark,coltitle=white,title={AIME 2026: Objective}]
Improve exact-match accuracy on AIME 2026 under the same per-task budget while retaining other capabilities. Aggregate benchmark measurements are optional and count toward the round budget.
\end{tcolorbox}